\documentclass[10pt,twocolumn,letterpaper]{article}

\usepackage[pagenumbers]{cvpr} % To force page numbers, e.g. for an arXiv version

\usepackage{graphicx}
\usepackage{amsmath}
\usepackage{amssymb}
\usepackage{booktabs}

\newcommand{\reffig}[1]{Fig.~\ref{#1}}
\newcommand{\refeq}[1]{Eq.~\ref{#1}}
\newcommand{\reftable}[1]{Table.~\ref{#1}}
\newcommand{\refsec}[1]{Sec.~\ref{#1}}

\usepackage[pagebackref,breaklinks,colorlinks]{hyperref}

\usepackage[capitalize]{cleveref}
\crefname{section}{Sec.}{Secs.}
\Crefname{section}{Section}{Sections}
\Crefname{table}{Table}{Tables}
\crefname{table}{Tab.}{Tabs.}

\def\cvprPaperID{10091}
\def\confName{CVPR}
\def\confYear{2022}

\begin{document}

\title{MDAN: Multi-level Dependent Attention Network for Visual Emotion Analysis}
% \vspace{0.25cm}
\author{Liwen Xu \hspace{0.8em} Zhengtao Wang \hspace{0.8em} Bin Wu \hspace{0.8em} Simon Lui\\
% \vspace{0.25cm}
Innovative Technology Center, Tencent Music Entertainment\\
Shenzhen, China\\
{\tt\small s1052262195@gmail.com, moyanwang@tencent.com, benbinwu@tencent.com, nomislui@gmail.com}
}

\maketitle

%%%%%%%%% ABSTRACT
\begin{abstract}
Visual Emotion Analysis (VEA) is attracting increasing attention.
One of the biggest challenges of VEA is to bridge the affective gap between visual clues in a picture and the emotion expressed by the picture.
As the granularity of emotions increases, the affective gap increases as well.
Existing deep approaches try to bridge the gap by directly learning discrimination among emotions globally in one shot. 
They ignore the hierarchical relationship among emotions at different affective levels, and the variation in the affective level of emotions to be classified.
In this paper, we present the multi-level dependent attention network (MDAN) with two branches to leverage the emotion hierarchy and the correlation between different affective levels and semantic levels.
The bottom-up branch directly learns emotions at the highest affective level and
largely prevents hierarchy violation by explicitly following the emotion hierarchy while predicting emotions at lower affective levels. 
In contrast, the top-down branch aims to disentangle the affective gap by one-to-one mapping between semantic levels and affective levels, namely, Affective Semantic Mapping.
A local classifier is appended at each semantic level to learn discrimination among emotions at the corresponding affective level.
Then, we integrate global learning and local learning into a unified deep framework and optimize it simultaneously.
Moreover, to properly model channel dependencies and spatial attention while disentangling the affective gap, we carefully designed two attention modules: the Multi-head Cross Channel Attention module and the Level-dependent Class Activation Map module.
Finally, the proposed deep framework obtains new state-of-the-art performance on six VEA benchmarks, where it outperforms existing state-of-the-art methods by a large margin, e.g., +3.85\% on the WEBEmo dataset at 25 classes classification accuracy.

\end{abstract}

\section{Introduction}
Visual Emotion Analysis (VEA) is a high-level abstraction task, which aims to recognize the emotion induced via visual content. Recently, it raises more and more research attention due to the trend that social network users become more likely to express their opinions and emotions via visual content. VEA has many practical applications, such as opinion mining \cite{You2016CrossmodalityCR}, business intelligence, entertainment assistant \cite{Zhao2021AffectiveIC} and personalized emotion prediction \cite{Zhao2016PredictingPE}.

One of the biggest challenges of VEA is to bridge the \textit{affective gap} between pixel-level information and the high-level emotion semantics \cite{Machajdik2010AffectiveIC,zhao2014exploring,Zhao2021AffectiveIC}.
\begin{figure}[t] 
\centering
\includegraphics[width=\columnwidth]{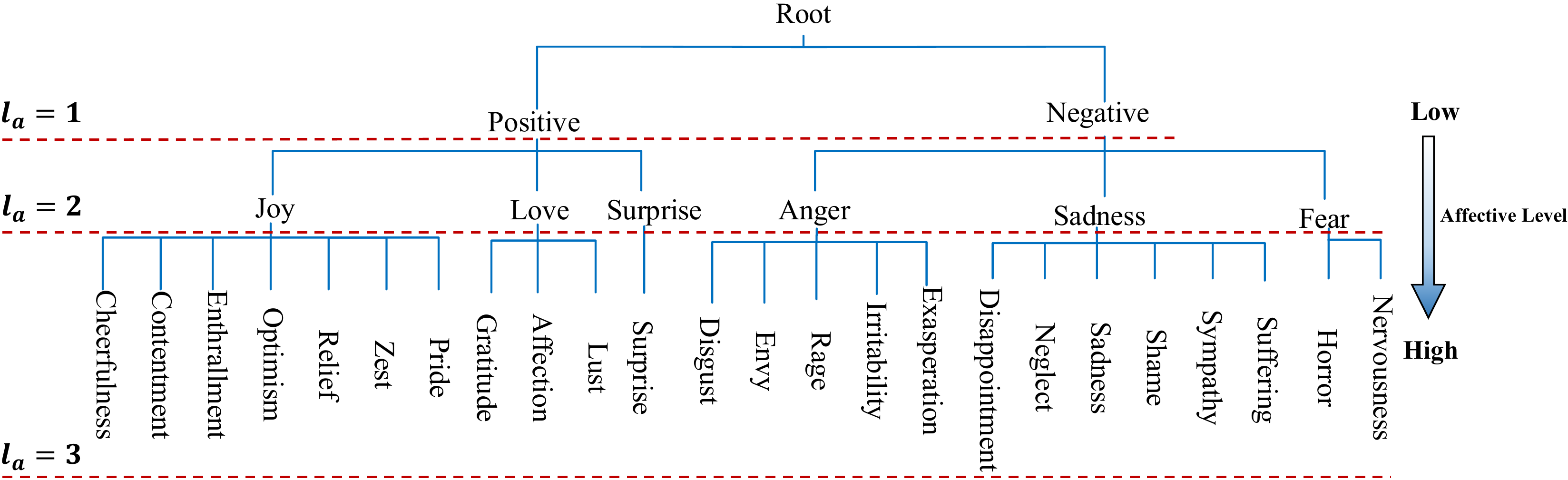}
\caption{Illustration of a three-level emotion hierarchy defined by Parrott in psychological studies. }
\label{emhierarchy}
\end{figure}
To bridge the affective gap, the key is to extract discriminative features \cite{Zhao2019PDANetPD}.
Recent efforts have been devoted to improving the discrimination of learned features for fine-grained emotion classification in one shot globally, i.e., learning one discriminative feature through one global classifier\cite{rao2017dependency,Zhao2019PDANetPD,She2020WSCNetWS,Zhang2020WeaklySE}.

However, as the granularity of emotions increases, the affective gap becomes larger because of the higher affective level, described by the arrow in \reffig{emhierarchy}, making the global learning in one shot difficult and unreliable.
In this work, we try to disentangle the affective gap into several smaller steps.
Instead of merely learning the discriminative feature for fine-grained emotions globally, we introduce an extra top-down branch to learn level-wise discrimination.
Specifically, we append a local classifier at each semantic level of the top-down branch and each of the local classifiers concentrates on learning the discrimination among emotions at a particular affective level.
\begin{figure}[ht!]
\centering
\includegraphics[width=\columnwidth]{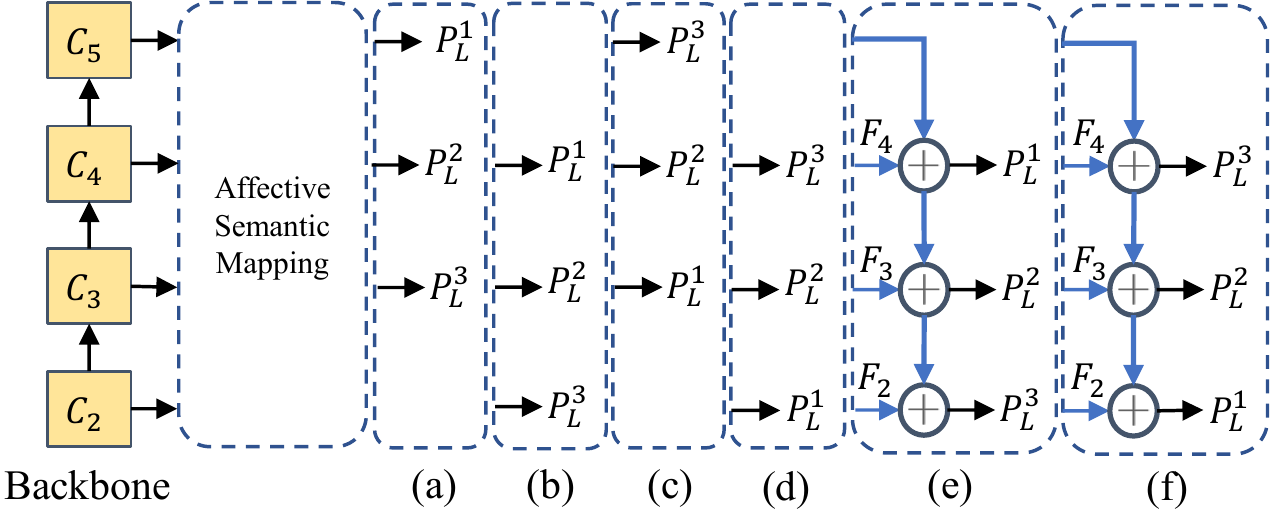}
\caption{Different mapping relationships between semantic levels of feature maps and affective levels of emotion categories with or without feature fusion.}
\label{fig:asm}
\end{figure}
In this paper, the term `local' is equivalent to `level-wise', since both of them refer to a particular affective level in an emotion hierarchy.
The above-mentioned setting for the top-down branch is based on an assumption that there is a correlation between different semantic levels and affective levels, namely, \textit{Affective Semantic Mapping}.
Feature maps extracted from CNN can have different semantic levels. While feature maps at lower semantic levels are spatially fine but semantically weak, those at higher semantic levels are spatially coarse but semantically strong \cite{Lin2017FeaturePN}.
Properly leveraging the trade-off between spatial richness and semantic strongness for classifying emotions at different affective levels is the key idea of Affective Semantic Mapping.
To verify the assumption, we examined a series of mapping settings shown in \reffig{fig:asm}.

Furthermore, most attention-based VEA methods extract one attended feature either based on the final feature map computed from the backbone \cite{Zhao2019PDANetPD,She2020WSCNetWS} or by directly concatenating multi-level features into one \cite{Zhang2020WeaklySE} without disentangling the multi-level information in a level-wise manner.
On the one hand, only considering the final feature map ignores low-level features, which are crucial for emotions classification \cite{lu2012shape,zhao2014exploring,Sartori2015WhosAO}, and can only perceive spatial features at very a limited scale.
On the other hand, trivial concatenation of multi-level feature maps may lead to negative interference between feature maps from different semantic levels.
As a result, the capability of existing methods for classifying fine-grained emotions is limited.

To address the above problems, we purpose a novel deep framework based on Feature Pyramid Network (FPN) \cite{Lin2017FeaturePN}, called as multi-level dependent attention network (MDAN), for fine-grained emotion classification, which is illustrated in \reffig{oa}.
It consists of two branches. The bottom-up branch classifies emotions at the highest affective level globally and largely prevents hierarchy violations by explicitly following emotion hierarchies.
The top-down branch disentangles the affective gap through several local classifiers. It focuses on learning discriminative representations at each semantic level for emotions at the corresponding affective level.
The two branches are optimized simultaneously.
Moreover, to leverage channel dependencies and spatial attention while disentangling the affective gap, we introduce two new attention modules, named Multi-head Cross Channel Attention (MHCCA) and Level-dependent Class Activation Map (L-CAM).
They are appended at each semantic level of the top-down branch.
For MHCCA, we refine and abridge multi-head attention mechanism \cite{Transformer}, and extend it from modeling pixel interdependencies \cite{Wang2018NonlocalNN,Carion2020EndtoEndOD,Dosovitskiy2021AnII,Srinivas2021BottleneckTF,Strudel2021SegmenterTF} into exploring channel attribute dependencies between feature maps at adjacent levels, \textit{Cross Attention}.
For L-CAM, it is similar to CAM proposed in \cite{Zhou2016LearningDF} but we leverage the subordinate relationship between emotions at adjacent affective levels. Specifically, the computed attention map is subject to the prediction results from the former affective level.

Our contributions are summarized as follow:
First, we provide a new perspective for VEA, which is to disentangle the affective gap by learning the level-wise discrimination. We study the Affective Semantic Mapping to achieve this.
Second, We purpose a novel multi-level dependent attention network consisting of two branches, classifying emotions both globally and locally in a simultaneous manner. Two new attention modules: the MHCCA module and the L-CAM module are appended at each semantic level to select important channel attributes and to highlight spatial details for each affective level.
Finally, the proposed deep framework obtains new state-of-the-art performance on six VEA benchmarks, where it outperforms existing state-of-the-art methods by a large margin, e.g., +3.85\% on WEBEmo, and +2.07\% on Emotion-6 at classification accuracy.

\section{Related Work}
\subsection{Emotion Models from Psychology}
Psychologists mainly employ two kinds of emotion representation models to measure emotion: categorical emotion states (CES) and dimensional emotion space (DES) \cite{Zhao2021AffectiveIC}.
CES models classify emotions into a few basic categories while DES models employ continuous 2D, 3D, or higher dimensional Cartesian space to represent emotions, such as valence-arousal-dominance (VAD).
In this paper, we mainly focus on CES for its superiority in understandability and popularity compared with DES.
The simplest CES model is binary \textit{positive} and \textit{negative}, which is too coarse-grained.
Some relatively fine-grained emotion models are designed, such as Ekman’s six emotions (anger, disgust, fear, happiness, sadness, surprise) \cite{ekman} and Mikels’s eight emotions (amusement, anger, awe, contentment, disgust, excitement, fear, and sadness) \cite{Mikels2005EmotionalCD}. These CES models can be expressed in a two-level tree hierarchical grouping.
With the development of psychological theories, categorical emotions are becoming increasingly diverse and fine-grained.
Parrott \cite{parrott2001emotions} proposed a three-level tree hierarchical grouping, which represents emotions with primary, secondary, and tertiary categories, illustrated in \reffig{emhierarchy}.
To fairly examine the improvement brought by invoking emotion hierarchy information, we selected six benchmark datasets that covered all aforementioned CES models.

\begin{figure}[ht!]
    \includegraphics[width=\columnwidth]{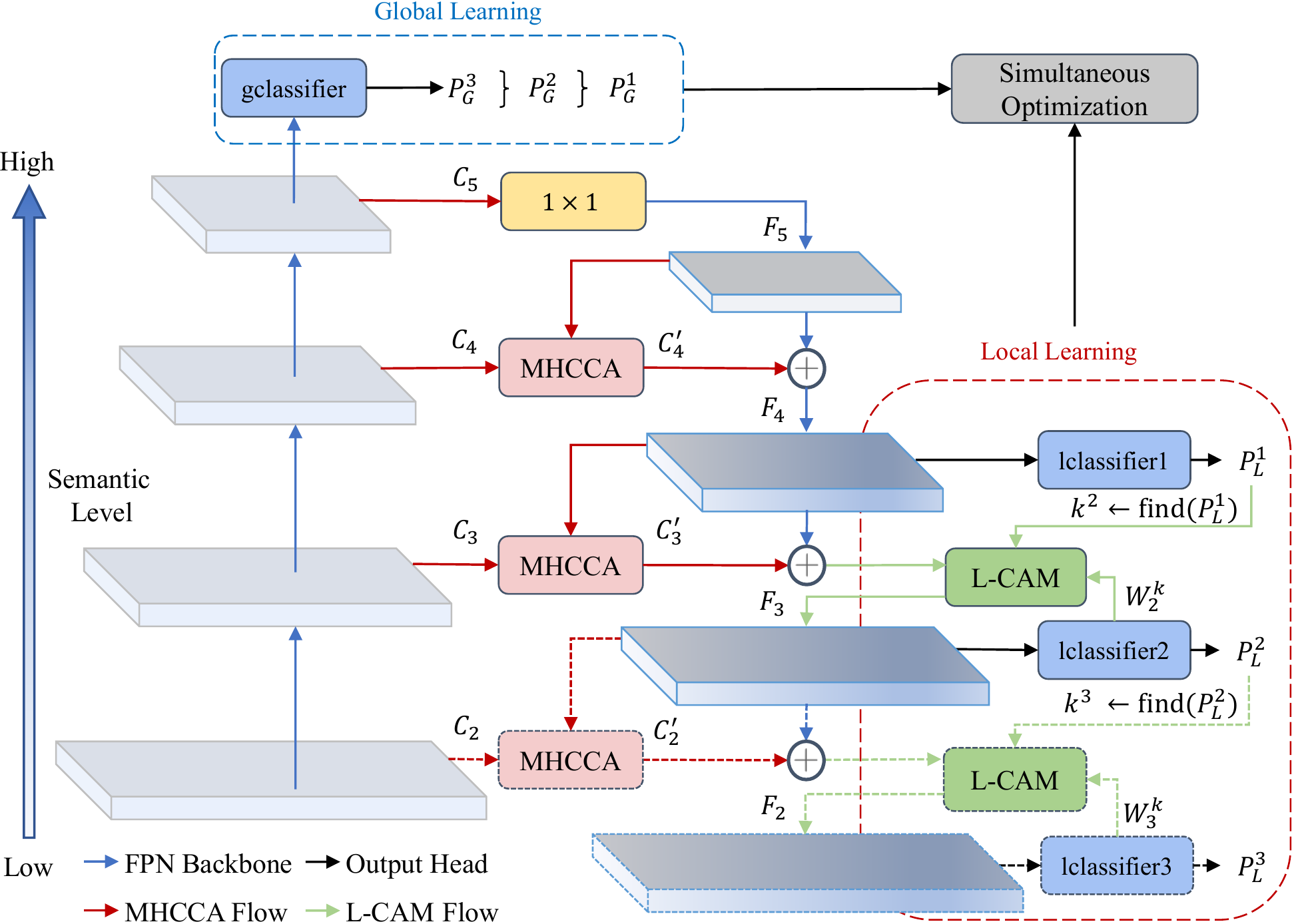}
    \caption{Overview of the multi-level dependent attention network. It consists of a bottom-up branch for global learning and a top-down branch for disentangling the affective gap and learn the level-wise discrimination. Components in dotted lines exist only when a three-level emotion hierarchy is used.}
    \label{oa}
\end{figure}

\subsection{Visual Emotion Analysis}
Existing attention-based deep approaches for VEA can be summarized as ``learning one discriminative feature through one global classifier''.
First, \cite{You2017VisualSA,Sun2016DiscoveringAR,yang2018auto} discover affective regions and produce attended feature maps based on a hard spatial attention map produced by an object proposal algorithm, after which the attended feature map is sent to a global classifier for emotion classification.
However, these two-stage methods are computationally expensive and are sub-optimal. This is because object detection is separated from the emotion prediction process, and regions that have no salient object may be excluded by the hard attention map.
Later, \cite{Zhao2019PDANetPD,She2020WSCNetWS} introduced a weakly supervised soft attention mechanism and employed it on the final feature map at the highest semantic level from the backbone.
\cite{Zhang2020WeaklySE} is most related to our work, as it is the only soft attention-based VEA method that considered multi-level information. The proposed emotion intensity map is learned by directly concatenating multi-level feature maps into one for learning the emotion intensity map.
Although these soft attention-based methods further strengthen the capability of modeling the discrimination among fine-grained emotions by leveraging spatially localized information, most of them either merely use the final feature map for classification or trivially concatenate multi-level feature maps into one to learn the discrimination among emotions in one shot globally without disentangling of level-wise discrimination among emotions at different affective levels.
Different from the above approaches, we try to disentangle the affective gap into several adjacent affective levels according to the emotion hierarchy and to bridge it level-wisely.

\subsection{Attention Mechanism}
Multi-head Self-Attention (MHSA) is firstly proposed in \cite{Transformer} for NLP tasks. It can not only model long-range dependencies explicitly but also jointly attend to information from different representation subspaces.
Recently, MHSA is increasingly applied in the image vision field. The pioneer work \cite{Wang2018NonlocalNN} which is highly related to SA, mainly explores the effectiveness of the non-local operation in space-time dimension for videos and images.
Later, the work \cite{Carion2020EndtoEndOD,Dosovitskiy2021AnII,Srinivas2021BottleneckTF,Strudel2021SegmenterTF} introduces MHSA on the spatial dimension to improve model's performance on object detection, object classification and semantic segmentation.
Different from previous works, we refine and abridge the MHSA module, and extend it to modeling attribute dependencies between feature maps at adjacent semantic levels on the channel dimension in the task of emotion classification. As level-wise discrimination propagate downward on the top-down branch, the proposed attention module, MHCCA, helps to highlight important channel attributes while suppressing redundancies. Ablation studies are conducted to verify the effectiveness of MHCCA.

\section{Affective Gap Disentanglement}\label{agd}

In this section, we will describe  the affective semantic mapping and the baseline version of MDAN without attention mechanisms.
Then, we will describe how local learning and global learning are simultaneously optimized.

\subsection{Affective Semantic Mapping}\label{asm}
Studies about the correlation between affective levels and semantic levels are not uncommon before the deep learning era. The works \cite{lu2012shape,zhao2014exploring,Sartori2015WhosAO} conclude that while emotions at high affective levels depend more on high-level semantic information, emotions at low affective levels is more relevant to low-level features like color, texture, \textit{etc}.
The conclusion is still valid in the early stage of deep learning. The mapping settings (a)-(d) in \reffig{fig:asm} we examined still follow the conclusion.
However, with the existence of feature fusion operation proposed in FPN \cite{Lin2017FeaturePN}, low-level feature maps can have both spatial richness and semantic strongness. This case is not covered in the aforementioned studies and hence the conclusion may not be applicable for this case.
We then assume that semantically strong and spatially rich feature maps at low semantic levels on the top-down branch of FPN may be more discriminative for classifying emotions at higher affective levels.
After examining the mapping settings (e)-(f), the results violate the above conclusion and verify our assumption.

Here we empirically choose (e) in \reffig{fig:asm} based on the discussion in \refsec{exp:asm}. Formally, $l_s = 5-l_a$ for $l_a \in {1,2,3}$, where $l_s$ is the semantic level and $l_a$ is the affective level.
An input image is first fed into the ResNet-101 and let $\{C_2, C_3, C_4, C_5\}$ as the feature maps produced by the last convolutional layer of the four bottleneck blocks on the bottom-up branch, respectively. 
They follow the trade-off between semantic strongness and spatial richness \cite{Lin2017FeaturePN} with $C_5$ as the semantically highest feature map and $C_2$ as the lowest one, shown by the arrow in \reffig{oa}.
Afterward, $C_5$ is fed to a pointwise convolution and its channel number is reduced, resulting in $F_5$. 
Then this semantically strong feature at a high level is passed forward and enrich the semantic information of low-level features by the following: $\rm \mathit{F}_{\mathit{l_s}} = Conv_{1\times1}(\mathit{C}_{\mathit{l_s}} )+Upsample_{2\times2}(\mathit{F}_{\mathit{l_s}+1})$
where $\rm Conv_{1\times1}$ denotes a pointwise convolution, $\rm Upsample_{2\times2}$ denotes bilinear upsampling by a factor of 2 and $l_s \in 2,3,4$. Because of the feature fusion, feature maps at a low semantic level have not only rich spatial details but also strong semantic meaning.
Afterward, $F_{l_s}$ is sent to the corresponding local classifier for level-wise learning and level-wise prediction.
Formally, we rely on several local classifiers and each of them independently learns the level-wise discrimination and makes local predictions. Specifically,
\begin{equation}
    \label{locallearning}
        \rm \mathit{P}_\mathit{L}^\mathit{l_a} = softmax(lclassifier_{\mathit{l_a=5-l_s}}(GAP(\mathit{F}_{\mathit{l_s}})))
\end{equation}
where $\rm GAP$ denotes global average pooling, $P_{L}^{l_a}$ is the local prediction at $l_a$ and $\rm lclassifier_{\mathit{l_a=5-l_s}}$ denotes the local classifier learning the discrimination at semantic level $l_s$ for emotions at affective level $l_a$.

\subsection{Global and Local Learning}\label{gll}
In \refsec{asm}, the local learning we described can, on the one hand, concentrate on a particular level and can better extract level-wise discrimination, which is crucial for fine-grained classification.
On the other hand, it is easy for local learning to overfit, which results in hierarchy violations among local predictions at different affective levels, \textit{i.e.}, an image is classified in \textit{positive} at $l_a=1$ but is classified in \textit{sadness} at $l_a=2$.
In contrast, global learning largely prevents hierarchy violations by explicitly following emotion hierarchies shown in \reffig{emhierarchy}.
However, it is likely to suffer from error propagation and be incapable of grasping the subtle discrimination among fine-grained categories. 
Optimizing the two objectives simultaneously helps to combine the advantages of both, and to avoid their drawbacks.

For global learning, a global classifier is appended at the top of the backbone to learn the discrimination based on $C_5$.
Specifically, $\rm \mathit{P_{G}^{|l_a|}} = softmax(gclassifier(GAP(\mathit{C}_5)))$,
where $\rm gclassifier$ denotes the global classifier and $|l_a|$ refers to the highest affective level under the used emotion hierarchy.
Then the global prediction for category $j$ at $l_a-1$, $P_{G,j}^{l_a-1}$ is acquired by summing the global probability of all children categories $k$ at $l_a$ of class $j$ at $l_a-1$. We used $\mathbf{\}}$ to represent this operation in \reffig{oa}. 
Formally,
\begin{equation}
    \label{childrensum}
    P_{G,j}^{l_a-1} = \sum_{k \in j} P_{G,k}^{l_a}
\end{equation}
where $P_{G,j}^{l_a-1}$ is the global prediction for class $j$ at $l_a-1$ and $k$ is the children categories of $j$.
In this way, hierarchy violations are largely avoided in the global prediction result.
For local learning, we have already discussed it in \refsec{asm}.

After getting the predictions from both global learning and local learning, they are weighted summed and are optimized simultaneously. Specifically, $P_{O}^{l_a} = \alpha \times P_{L}^{l_a} + (1 - \alpha) \times P_{G}^{l_a}$, 
where $P_{O}^{l_a}$ is the overall prediction at $l_a$. Except for $|l_a|$, global predictions at lower levels are calculated by \refeq{childrensum}. $\alpha$ is a hyper-parameter discussed in \refsec{exp:alpha}.
Since categories are mutually exclusive, we minimize the multi-class cross-entropy loss to learn both global and level-wise discrimination. Specifically,
\begin{equation}
    \label{loss}
        \mathcal{L}=\frac{1}{|l_a|}\sum_{i=1}^{|l_a|} \mathcal{L}_{i} = -\frac{1}{|l_a|}\frac{1}{N}\sum_{i=1}^{|l_a|}\sum_{j=1}^{N}\sum_{k=1}^{|C_{i}|} Y^i_{j,k} \times P^i_{O, j,k}
\end{equation}
where $N$ is the number of samples within a mini-batch, $|C_{i}|$ is the number of emotion categories at $l_a=i$ and $Y^i_{j,k}$ is the ground truth of sample $j$ for category $k$ at $l_a$.

\section{Attention Mechanism}
In this section, we will describe the details of the proposed two attention modules: MHCCA and L-CAM. They are appended at each semantic level of MDAN for modeling channel dependencies and spatial attention level-wisely.

\subsection{Multi-head Cross Channel Attention}\label{method:mhcca}
\begin{figure}[t]
\centering
\includegraphics[width=0.8\columnwidth]{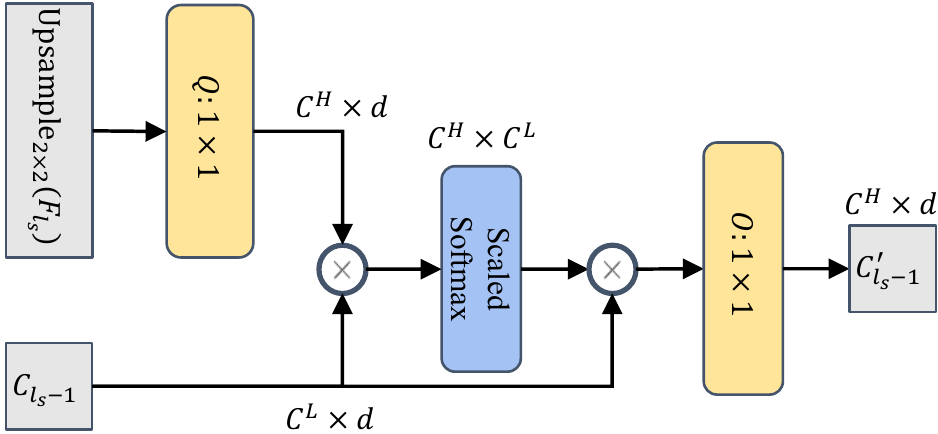}
\caption{Multi-head Cross Channel Attention employed in MDAN. While we use multiple heads, we do not show them on the figure for simplicity. $C^H$ refers to the number of channel attributes in $F_{l_s}$ and $C^L$ refers to that in $\mathit{C}_{\mathit{l_s}-1}$.}
\label{fig:mhcca}
\end{figure}

Each channel attribute can be regarded as a class-specific response, and different semantics are associated with each other.
Properly selecting important channel attributes is crucial for fine-grained emotion classification.
The structure of MHCCA is illustrated in \reffig{fig:mhcca}. It computes Cross-Attention (CA). CA attempts to discover the pair-wise channel dependencies between two feature maps at two adjacent semantic levels.
Specifically,
\begin{equation}
    \label{eq:mhcca}
        C_{l_s-1}^{\prime}=O_{1\times1}\left(\operatorname{softmax}\left(\frac{Q_{1\times1}\left(F_{l_s} \right) {C_{l_s-1}}^{T}}{\sqrt{d_{C_{l_s-1}}}}\right) C_{l_s-1} \right)
\end{equation}
where $Q_{1\times1}$ and $O_{1\times1}$ are pointwise convolution layers, $d_{C_{l_s-1}}$ is the flattened spatial dimension per head in $C_{l_s-1}$.
The above computation flow is illustrated in \reffig{fig:mhcca}.
By modeling CA, instead of trivial adding in FPN, channel attributes in $C_{l_s}$ that has stronger correlations with those in $\rm \mathit{F}_{\mathit{l_s}+1}$ are highlighted, and those less correlated are suppressed, leading to more discriminative features. 
Meanwhile, since classifying emotions at lower affective levels are comparatively easier \cite{Panda2018ContemplatingVE}, it is more likely to learn robust and reliable level-wise discrimination at lower affective levels.
As the affective level go up, classification becomes increasingly difficult. Reliable level-wise discrimination at $l_a$ could guide the model to concentrate on important channel attributes and avoid learning degradation at $\rm \mathit{l}_{\mathit{a}}+1$.

\subsection{Level-dependent Class Activation Map}
\subsubsection{Class Activation Map} Apart from channel interdependencies, spatial attention is also essential, as it highlights the regions of interest that are informative for emotion classification \cite{Zhao2019PDANetPD,She2020WSCNetWS}. 
In particular, we use the work of \cite{Zhou2016LearningDF} to generate the class activation map (CAM).
A CAM is a topographic heat map that illustrates which parts of the image are important for the classification of a particular class.
As described in \refeq{locallearning}, at each $l_a$, we have a $\rm lclassifier_\mathit{l_a}$ to learn the level-wise discrimination. 
Based on this design, we can compute a CAM at each affective level. 
Let the weight vector in $\rm lclassifier_\mathit{l_a}$ for class $k$ is $w^k_{l_a}$ and the class activation map for class $k$ at $l_a$ be denoted as $M^{k}_{l_a}$, then the map can be computed as $M^{k}_{l_a}(x,y)=\sum_c w^{k}_{l_a}F_{l_s = 5-l_a}^{c}(x,y)$, where $F_{5-l_a}^{c}$ denotes the $c^{th}$ channel attribute in $F_{5-l_a}$.

\subsubsection{L-CAM}
\begin{figure}[t]\small
\centering
\includegraphics[width=0.78\columnwidth]{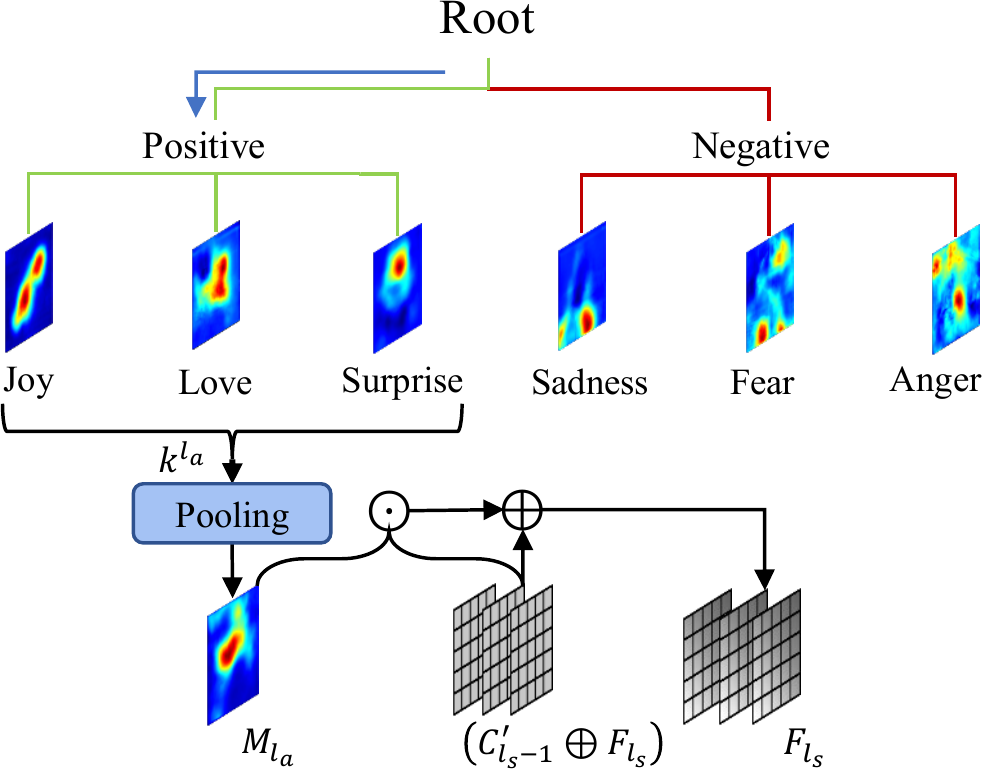}
\caption{Illustration of the computation flow of L-CAM and the $\rm \mathit{k}^{\mathit{l_a}} \leftarrow find(\mathit{P_L}^{\mathit{l_a}-1})$ operation in \reffig{oa}.}
\label{lcam}
\end{figure}

CAM cannot be used directly in a forward propagation with only one classifier.
However, since multiple local classifiers learn different level-wise discrimination, we can leverage the subordinate relationship between adjacent levels in the emotion hierarchy and highlight spatial positions that are informative for classifying at $l_a$ and are consistent with its ancestor at $\rm \mathit{l}_\mathit{a}-1$.
To achieve this, we purpose L-CAM, as shown in \reffig{lcam}.
In particular, the discriminative regions that are to be highlighted at $l_a$ are subject to the local prediction from the former affective level $P_{L}^{l_a-1}$. 
For example, in \reffig{lcam}, $P_{L}^{1}$ is \textit{positive} and, consequently, the CAM of children categories of \textit{positive}, $\rm \{joy,love,surprise\}$ are computed to guide the learning and predicting at $l_a=2$, i.e. $P_{L}^{2}$. Formally,
$M_{l_a}(x,y) = \frac{1}{|k|}\sum_{k \in j}M_{l_a}^{{k}}(x,y) + \max_{k \in j}M_{l_a}^{{k}}(x,y)$
where $k$ is the children categories of $j$. 
Then, they are fused by pooling operations into one informative spatial attention map, guiding the classification for emotions at higher affective level. Here we use $mean$ and $max$ as pooling operations along the spatial dimension. L-CAM takes $(C^\prime_{l_s-1} \oplus F_{l_s})$ as the input, and the attended output is computed by $F_{l_s} = (1 + M_{l_a}) \odot (C^\prime_{l_s-1} \oplus F_{l_s})$, where $\odot$ refers to Hadamard Product by broadcasting $M_{l_a}$.

\section{Experiment}

\begin{table*}[t]\small
\renewcommand\arraystretch{0.55}
\resizebox{\linewidth}{!}{
\begin{tabular}{l|c|c|cc|cc|ccc|cc|cc|cc}\toprule
Model & \#params & GFLOPs & \multicolumn{2}{c}{Emotion} & \multicolumn{2}{c}{FI} & \multicolumn{3}{c}{WEBEmo} & \multicolumn{2}{c}{Artphoto} & \multicolumn{2}{c}{IAPSsubset} & \multicolumn{2}{c}{Abstract} \\\midrule
\multicolumn{3}{l}{Number of classes} & 2 & 6 & 2 & 8 & 2 & 6 & 25 & 2 & 8 & 2 & 8 & 2 & 8 \\\midrule
Sentibank \cite{Borth2013SentiBankLO} & - & - & - & - & 56.47 & 44.49 & - & - & - & 67.74 & 53.96 & 81.79 & 73.58 & 64.95 & 50.68 \\
DeepSentibank \cite{Chen2014DeepSentiBankVS} & - & - & - & - & 64.39 & 53.16 & - & - & - & 68.73 & 68.54 & 85.63 & 75.88 & 71.19 & 66.46 \\
AlexNet \cite{Krizhevsky2012ImageNetCW} & 61.10M & 2.8 & 75.88 & 51.35 & 72.43 & 58.30 & - & 48.66 & 28.21 & 69.27 & 67.03 & 84.58 & 72.24 & 65.49 & 61.96 \\
VGGNet-16 \cite{Simonyan2015VeryDC} & 138M & 62.0 & - & - & 83.05 & - & - & - & - & 70.09 & 68.16 & 88.51 & 74.78 & 72.48 & 62.41 \\
PCNN \cite{You2015RobustIS} & - & - & - & - & 75.34 & 56.16 & - & - & - & 70.96 & 68.93 & 88.84 & 76.87 & 70.84 & 67.17 \\
ResNet-50 \cite{He2016DeepRL} & 23.52M & 15.6 & 79.23 & 54.99 & 85.43 & 64.74 & 76.65 & 51.97 & 31.80 & 70.93 & - & 89.95 & - & 73.07 & - \\
ResNet-101 \cite{He2016DeepRL} & 42.52M & 30.4 & 79.78 & 56.69 & 85.92 & 66.16 & 78.17 & 52.35 & 32.14 & 71.08 & 69.36 & 90.13 & 75.09 & 73.36 & 63.56 \\\midrule
Zhu \textit{et al.} \cite{rao2017dependency} & - & - & - & - & 84.26 & 73.03 & - & - & - & 75.50 & 71.63 & 91.38 & 82.39 & 73.88 & 68.45 \\
Panda \cite{Panda2018ContemplatingVE} & 23.52M & - & 77.72 & - & 84.81 & - & 81.41 & - & - & - & - & - & - & - & - \\
Yang \textit{et al.} \cite{yang2018auto} & - & - & - & - & 86.35 & - & - & - & - & 74.80 & - & 92.39 & - & 76.03 & - \\
Rao \textit{et al.} \cite{rao2019multi} & 48.80M & - & 81.87 & - & 87.51 & 75.46 & - & - & - & 77.28 & 74.58 & 93.66 & 84.71 & 77.77 & 70.77 \\\midrule
WSCNet \cite{She2020WSCNetWS} & 42.60M & 62.7 & 82.15 & 58.47 & 86.74 & 70.07 & 79.43 & 52.61 & 32.75 & 80.38 & 72.86 & 94.61 & 82.25 & 77.84 & 64.45 \\
PDANet \cite{Zhao2019PDANetPD} & 63.51M & 67.6 & 82.27 & 59.24 & 87.25 & 72.13 & 80.96 & 53.46 & 32.82 & 80.27 & 74.62 & 95.18 & 80.92 & 78.24 & 67.13 \\
Zhang \cite{Zhang2020WeaklySE} & 50.29M & 128.3 & 82.95 & 60.41 & 90.97 & 75.91 & 82.47 & 53.88 & 33.01 & 79.24 & - & 95.83 & - & 83.02 & - \\
Ours & 48.79M & 101.4 & \textbf{84.62} & \textbf{61.66} & \textbf{91.08} & \textbf{76.41} & \textbf{82.72} & \textbf{55.65} & \textbf{34.28} & \textbf{91.50} & \textbf{78.12} & \textbf{96.73} & \textbf{85.96} & \textbf{83.80} & \textbf{72.34}\\\bottomrule
\end{tabular}
}
\caption{Model complexity, classification accuracy (\%) of the proposed framework for emotion categories at different $l_a$ for $l_a \in \{1,2,3\}$ on the testing set of Emotion6, FI, WEBEmo, ArtPhoto, IAPSsubset, Abstract datasets, and the comparison with previous work.}
\label{table:cp}
\end{table*}

\subsection{Dataset} \label{exp:dataset}
We evaluate our framework on six benchmark datasets annotated under different emotion hierarchies, namely, WEBEmo, Emotion-6 \cite{Panda2018ContemplatingVE}, the Flickr and Instagram (FI) \cite{You2016BuildingAL} dataset and the Artphoto, IAPS and Abstract dataset \cite{Machajdik2010AffectiveIC}.
WEBEmo is currently the largest dataset that consists of approximately 268,000 images retrieved from the Web with up to 25 emotion categories under Parrotts' three-level emotion hierarchy \cite{parrott2001emotions} as shown in \reffig{emhierarchy}. 
We use WEBEmo-25, WEBEmo-6, and WEBEmo-2 to denote annotations at different affective levels.
Emotion-6 dataset consists of 8350 and is annotated under Ekman's emotion hierarchy \cite{ekman}.
FI dataset consists of 23308 images and is annotated under Mikels' emotion hierarchy \cite{Mikels2005EmotionalCD}. 
Moreover, we also evaluate the proposed framework on three small-scale datasets, namely, the Artphoto, IAPS and Abstract datasets. They consist of 806, 395, 228 photos, respectively.

\subsection{Implementation Details} \label{exp:setting}
Our framework is implemented using PyTorch.
Our framework is based on the pretrained ResNet-101.
We train the models using stochastic gradient descent (SGD) for 30 epochs on two NVIDIA TESLA V100 GPUs with 64 GB onboard memory. Batch size, weight decay and momentum are set to 64, 0.001 and 0.9, respectively. 
The learning rates of the backbone and of newly defined layers are initialized to 0.001 and 0.01, respectively for large-scale datasets, and drop by a factor of 10 every 10 epochs.
For small-scale datasets, we first pretrain the model on FI dataset and then finetune it on small-scale datasets. 
The learning rates of the backbone and of newly defined layers are initialized to 0.0001 and 0.001, respectively. 
Following the configurations in \cite{Panda2018ContemplatingVE,Zhang2020WeaklySE}, the WEBEmo and the Emotion-6 dataset are split into 80\% and 20\% for training and testing, respectively. 
For the FI dataset, we split it into 80\% training, 5\% validation, and 15\% testing sets, to follow the configurations in \cite{You2016BuildingAL,She2020WSCNetWS}.
We run each of the models three times on Emotion-6 and FI datasets, and report the averaged result. We run once on WEBEmo for its large test set size.
For small scale datasets, \textit{i.e.,} Artphoto, IAPS and Abstract, we follow the setup in \cite{rao2017dependency,Zhang2020WeaklySE} and use 5-fold cross-validation and report the averaged result.
In addition, a $448 \times 448$ image is randomly cropped from each original image, and a horizontal flip is applied to the cropped image. 
Afterward, each channel of input data is normalized to have a zero mean and unit variance. 

\begin{table}[]
\renewcommand\arraystretch{0.6}
\resizebox{\columnwidth}{!}{
\begin{tabular}{@{}cc|ccc|cc|cc@{}}
\toprule
\multicolumn{2}{c|}{Mapping} & \multicolumn{3}{c|}{WEBEmo} & \multicolumn{2}{c|}{FI} & \multicolumn{2}{c}{Emotion} \\ \midrule
\multicolumn{2}{c}{Number of Classes} & 2 & 6 & 25 & 2 & 8 & 2 & 6 \\ \midrule
\multicolumn{1}{c|}{Baseline} & ResNet-101 & 78.17 & 52.35 & 32.14 & 85.92 & 66.16 & 79.78 & 56.69 \\ \midrule
\multicolumn{1}{c|}{w/o FF} & (a) & 78.01 & 47.64 & 25.69 & 87.30 & 67.52 & 80.31 & 57.00 \\
\multicolumn{1}{c|}{} & (b) & 77.94 & 46.91 & 23.17 & 84.79 & 64.80 & 77.48 & 54.41 \\
\multicolumn{1}{c|}{} & (c) & \textbf{78.71} & \textbf{51.35} & \textbf{31.23} & \textbf{88.27} & \textbf{69.84} & \textbf{81.53} & \textbf{58.64} \\
\multicolumn{1}{c|}{} & (d) & 78.13 & 49.12 & 30.81 & 85.03 & 67.29 & 77.51 & 56.95 \\ \midrule
\multicolumn{1}{c|}{with FF} & (e) & \textbf{79.79} & \textbf{52.11} & \textbf{31.85} & \textbf{90.36} & \textbf{70.55} & \textbf{81.94} & \textbf{59.62} \\
\multicolumn{1}{c|}{} & (f) & 79.72 & 51.74 & 31.68 & 89.51 & 70.46 & 81.63 & 59.00 \\ \bottomrule
\end{tabular}
}
\caption{Comparison of classification accuracy (\%) under different semantic affective mappings based on $P_L$ only. 'FF' is the abbreviation of 'Feature Fusion'.}
\label{table:asm}
\end{table}
\begin{table}[t]\scriptsize
    \renewcommand\arraystretch{0.55}
    \centering
    \resizebox{\columnwidth}{!}{
        \begin{tabular}{lcccc} \toprule
            Dataset & $\rm Accuracy_{L}$ & $\rm Accuracy_{G}$ & $\rm Accuracy_O$ \\\midrule
            Emotion-6 & 60.73 & 58.83 & \textbf{61.66}  \\
            FI-8 & 75.21 & 72.75 & \textbf{76.41} \\
            WEBEmo-6 & 53.88 & 53.34 & \textbf{55.65} \\
            WEBEmo-25 & 34.03 & 32.95 & \textbf{34.28} \\\midrule
            Emotion-2 & 84.44 & 81.48 & \textbf{84.62} \\
            FI-2 & 89.50 & 87.39 & \textbf{91.08} \\
            WEBEmo-2 & 82.53 & 79.91 & \textbf{82.72} \\
            Abstract-2 & \textbf{84.15}  & 80.64 & 83.80 \\
            IAPSsubset-2 & 95.75 & 95.25 & \textbf{96.73} \\
            ArtPhoto-2 & \textbf{92.50} & 89.38 & 91.50 \\
            \midrule
        \end{tabular}}
    \caption{Comparison of classification Accuracy (\%) calculated based on $P_L$, $P_G$ and $P_O$ on all datasets}
    \label{exp:gl}
\end{table}

\begin{figure*}[!t]
	\begin{center}
		\begin{tabular}{cc}
		\includegraphics[width=0.48\linewidth]{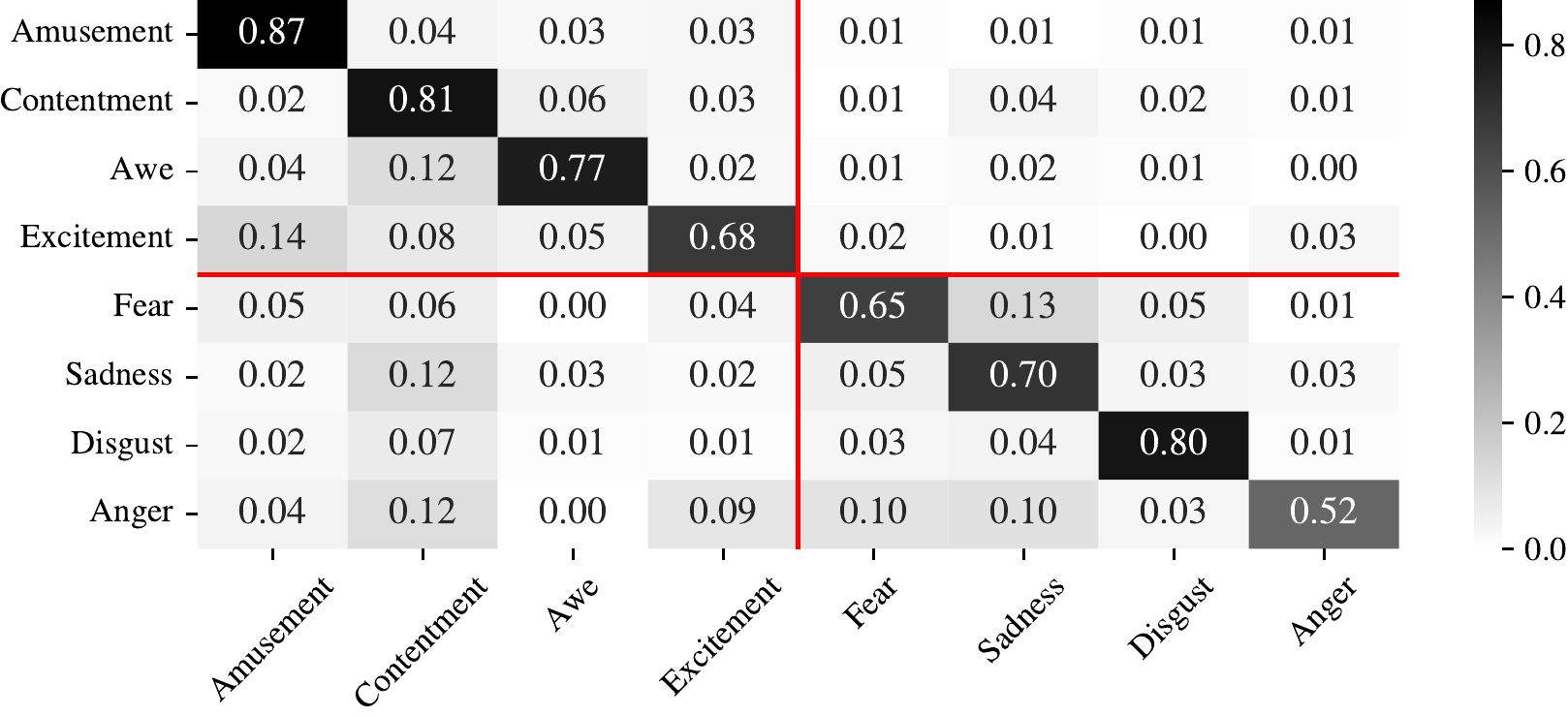} & \includegraphics[width=0.48\linewidth]{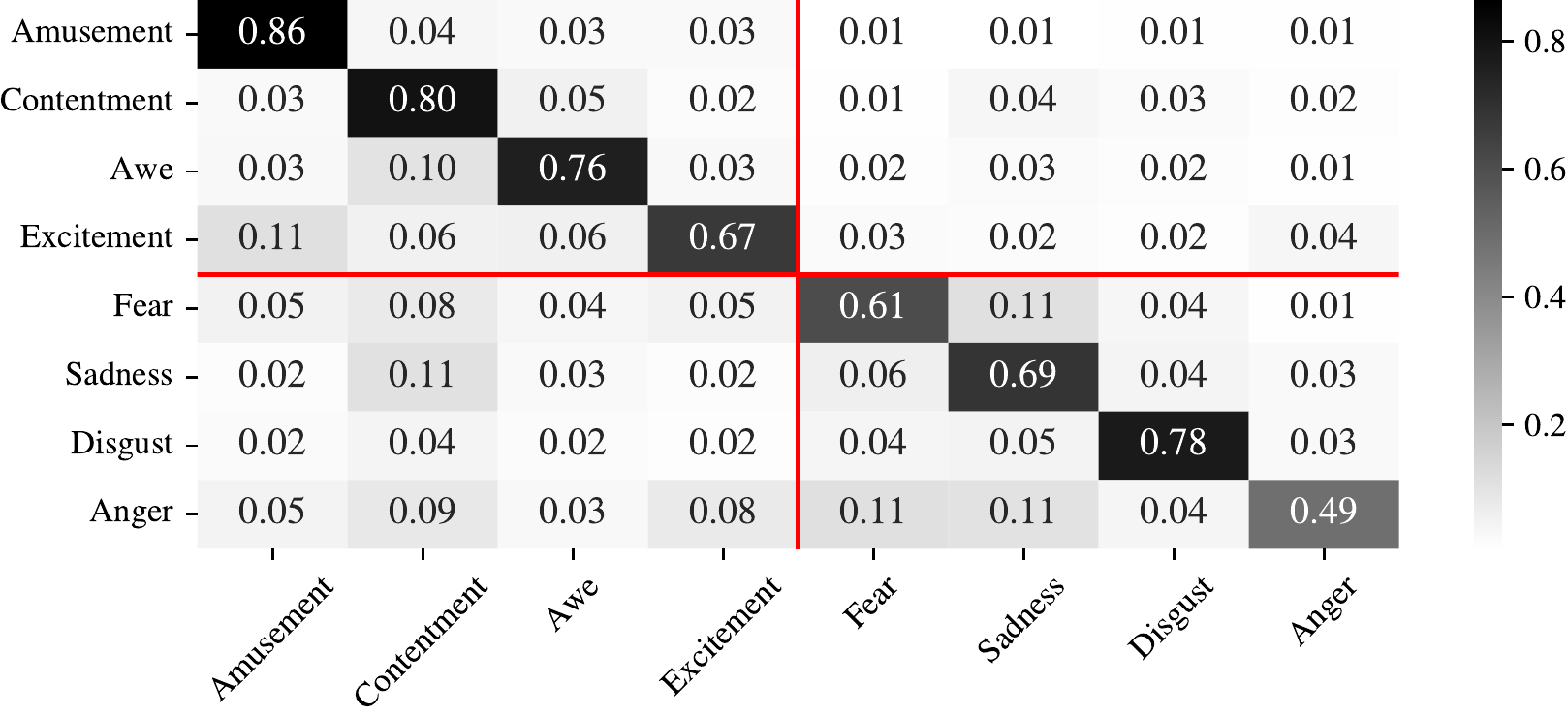}\\
		\end{tabular}
		\end{center}
		\vspace{-3mm}
		\caption{Confusion Matrix of overall prediction $P_O$ (left) and local prediction $P_L$ (right) on FI-8. The red lines divide emotions with a different parent classes. Wrong cases at top-right and bottom-left are in hierarchy violation. The grayscale is on the right.}
		\label{fig:cm}
\end{figure*}

\subsection{Affective Semantic Mapping} \label{exp:asm}

We first evaluate different mapping settings in \reffig{fig:asm} with local learning only based on the ResNet-101 and ResNet-101 based FPN backbone. 
\reftable{table:asm} shows the classification accuracy on the WEBEmo, FI and Emotion dataset under different mapping settings. 
The comparison between (a) and (b) shows that the classification accuracy of emotions at high affective levels, i.e., $P^2_L$ and $P^3_L$ are highly correlated with high semantic levels while emotions at low affective levels, i.e., $P^1_L$ is not significantly affected by semantic level.
This is consistent with the finding in \cite{lu2012shape,zhao2014exploring,Sartori2015WhosAO}, where a high-level semantic feature is more correlated to emotions at a high affective level while a low-level feature is more correlated to emotions at a low affective level.
Comparisons between (a) and (c), between (b) and (d), and between (e) and (f) indicate the importance of the order while disentangling the affective gap. 
Learning emotions at lower affective levels at first benefits the learning for emotions at higher affective levels. The affective gap has to be disentangled from low to high instead of high to low.
The three comparisons also verify our assumption that there is a correlation between different semantic levels and affective levels.
The comparison between (c) and (e) illustrates the existence of feature fusion improved the overall classification performance.
More importantly, feature fusion inverted the mapping relationship between semantic levels and affective levels mentioned in \cite{lu2012shape,zhao2014exploring,Sartori2015WhosAO}.
With feature fusion, the feature map at the lowest semantic level has both strong semantic meaning and spatial richness, resulting in a discriminative representation for classifying emotions at higher affective levels.
This comparison also verifies our assumption that the existence of feature fusion is not considered in the previous works and it inverted the mapping relationship between semantic levels and affective levels.

\subsection{Classification Performance} \label{exp:cp}

The experimental results of the proposed methods are compared with the following deep networks including
hand-crafted approaches Sentibank \cite{Borth2013SentiBankLO} 
, CNN-based approaches: \cite{Krizhevsky2012ImageNetCW,Chen2014DeepSentiBankVS,Simonyan2015VeryDC,You2015RobustIS,He2016DeepRL}
, multi-level based approaches,\cite{rao2017dependency,rao2019multi}
and recently proposed attention-based approaches: PDANet \cite{Zhao2019PDANetPD}, WSCNet \cite{She2020WSCNetWS}, Zhang \textit{et al.} \cite{Zhang2020WeaklySE}. 
\reftable{table:cp} shows that our proposed method consistently performs favorably against the state-of-the-art methods at all affective levels.
For the largest dataset WEBEmo, our approach outperforms current state-of-the-art by a large margin considering the large test set size of WEBEmo, 1.77\% and 1.27\%, on WEBEmo-6 and WEBEmo-25, respectively.
For small datasets, MDAN outperforms existing attention-based works by a large margin, especially on the Artphoto dataset, indicating the importance of disentangling attentions in a level-wise manner.
Moreover, we evaluated the effectiveness of simultaneously optimizing global learning $P_G$ and local learning $P_L$.
The confusion matrix on FI-8 obtained through overall prediction $P_O$ and local learning $P_L$ is shown in \reffig{fig:cm}. 
Compared with $P_L$, $P_O$ has fewer hierarchy violation cases by observing the top-right and bottom-left of confusion matrix above.
We further observe the accuracy gap among $P_G$, $P_L$, $P_O$ over all datasets and the results are shown in \reftable{exp:gl}.
Except for the results on Abstract and Artphoto, $P_O$ prevails $P_L$ and $P_G$ on most datasets.
This illustrates that by simultaneous optimization, MDAN could properly extract discriminative level-wise features through local learning and reduce the cases in hierarchy violation through global learning.
The two datasets consist of abstract paintings which lack salient objects or semantic meaning but full of low-level features such as texture, color, etc.
We argue that the $F_3$ that has not only rich semantic information but also rich in spatial details can be better used for classifying abstract pictures.
Overall, under a wider application scenario, our approaches can effectively reduce the hierarchy violation and improve the classification performance.

\begin{table}[!t]\small
    \renewcommand\arraystretch{0.5}
    \centering
    \resizebox{\columnwidth}{!}{
        \begin{tabular}{cccccccc}\toprule
            $h_{4}$ & $h_{3}$ & $d_{4}$ & $d_{3}$ & FI-2 & FI-8 & Emo-2 & Emo-6   \\\midrule
            1 &  1   &  784  &  3136   & 90.23 & 74.78 & 83.44 & 59.62 \\
            2 &  2   &  \textbf{392}  &  1568   & \textbf{90.83} & 75.85 & \textbf{84.37} & 60.75 \\ 
            4 &  4   &  196  &   \textbf{784}   & 90.31 & \textbf{76.24} & 84.35 & \textbf{61.36} \\
            8 &  8   &   98   &   392   & 89.37 & 76.01 & 83.62 & 60.53 \\
            16&  16  &   49   &   196   & 89.31 & 75.41 & 82.88 & 60.00 \\\midrule
            2 &  4   &  \textbf{392}   &   \textbf{784}   & \textbf{90.89} & \textbf{76.28} & 84.39 & \textbf{61.44} \\
            4 &  8   &  196   &   392   & 90.18 & 76.03 & \textbf{84.41} & 60.58 \\\bottomrule
        \end{tabular}}
    \caption{Effect of $h_{l_s}$ at different semantic levels. Comparison of dimension of subspaces $d_{l_s}$, and the corresponding classification accuracy (\%) on the FI and Emotion datasets.}
    \label{exp:heads}
\end{table}

\subsection{Hyper-parameter Analysis} \label{exp:ha}

\subsubsection{Number of attention heads}\label{heads}
The number of attention heads $h$ in MHCCA decides how many subspaces will a feature map be divided on the spatial dimension while computing the channel interdependencies. We set $\alpha=0.6$ while deciding the best setting for $h$.
\reftable{exp:heads} shows a jump in classification performance when $h$ doubled from 1 to 2, illustrating the effectiveness of spatially dividing each feature map into $h$ subspaces.
However, while single-head performs worse than the best setting, the performance also drops with too many heads, \textit{i.e.} 16.
Moreover, we found that different $h$ fit different semantic levels. At the high semantic level, the feature map already has a large respective field and thus we don't need a large spatial subspace size for modeling the channel dependencies. 
On the contrary, a low-level feature map has a relatively smaller respective field and a relatively larger subspace size is needed to properly capture the channel dependencies.
Therefore, we set $h_{l_s}=2,4,8$ for a trade-off between efficiency and effectiveness, where $l_s \in \{4,3,2\}$.

\begin{table}[!t]
    \renewcommand\arraystretch{0.6}
    \centering
        \resizebox{\columnwidth}{!}{
        \begin{tabular}{cccc|cc|c}\toprule
         $Base$ & MHCCA & $K_{1\times1}$ \& $V_{1\times1}$ & UpsampleAdd & $Mean$ & $Max$ & WEBEmo-25\\\midrule
         \checkmark & & & \checkmark & & & 32.78 \\
         \checkmark & \checkmark & \checkmark & \checkmark & & & 32.74 \\
         \checkmark & \checkmark & & \checkmark & & & 33.75 \\
         \checkmark & \checkmark &  &  & & & 33.13 \\
        %  \checkmark & \checkmark &  & \checkmark & & & 33.96 \\
         \midrule
         \checkmark & \checkmark &  & \checkmark & \checkmark &  & 34.04 \\
         \checkmark & \checkmark &  & \checkmark &  & \checkmark & 34.17 \\
         \checkmark & \checkmark &  & \checkmark & \checkmark & \checkmark &  34.28\\
         \checkmark &  &  & \checkmark & \checkmark & \checkmark &  33.48\\
         \bottomrule
        \end{tabular}
        }
    \caption{Ablation study on WEBEmo-25. $K_{1\times1}$ \& $V_{1\times1}$ denotes pointwise convolution layer for computing $K$ and $V$ from $\rm \mathit{C}_{\mathit{l_s}-1}$. $Mean$ and $Max$ represent the pooling operations in L-CAM.}
    \label{abstudy}
\end{table}

\begin{figure}[t]
\centering
\includegraphics[width=0.9\columnwidth]{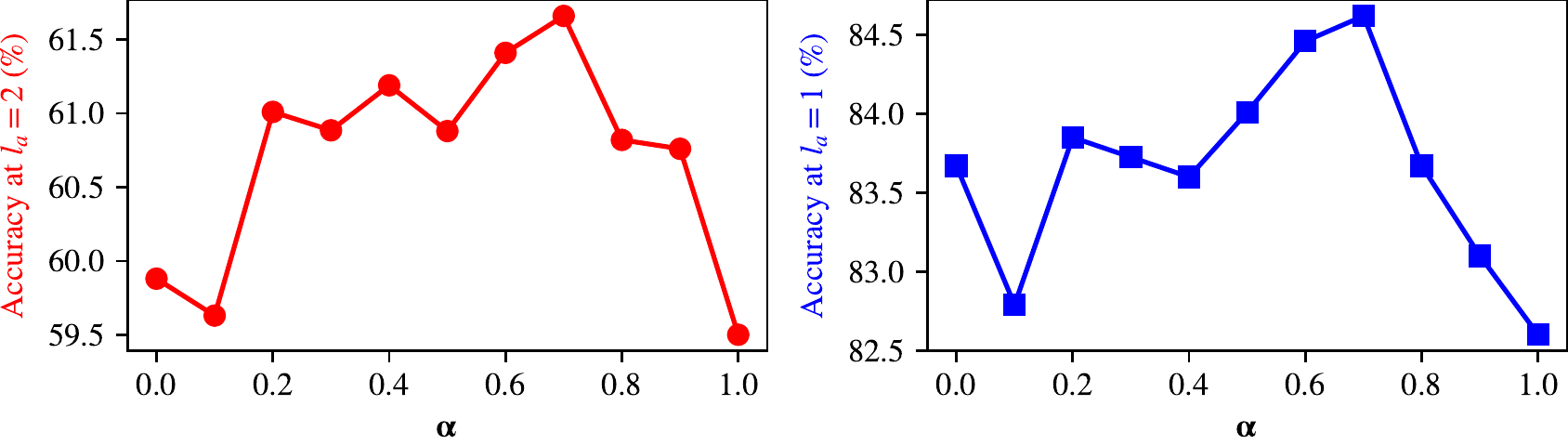}
\caption{Effect of $\alpha$ for fusing $P_L$ and $P_G$ on classification accuracy (\%) on Emotion-6 dataset. $\alpha$ is the weight of $P_L$.}
\label{fig:alpha}
\end{figure}

\subsubsection{Alpha}\label{exp:alpha}
$\alpha$ controls the relative importance between $P_L$ and $P_G$.
Based on Emotion-6 dataset, we discuss the influence of $\alpha$ on the classification performance at both $l_a=1$ and $l_a=2$. 
In \reffig{fig:alpha}, the accuracy variation at two levels are shown when $\alpha$ ranges from 0.0 to 1.0. 
While $0.0 \leq \alpha \leq 0.7$, the general trend rises although there are irregular bumps.
This indicates the complementarity between the information learned through global learning and local learning.
After achieving the peak at $\alpha=0.7$, the accuracy at both $l_a=1$ and $l_a=2$ drops sharply.
This may be caused by the dual effects of the over-fitting of local classifiers and the under-fitting of the backbone.

\subsection{Ablation Study}
We perform an ablation study to illustrate the effect of each contribution.
Our baseline is the FPN without attention mechanism but with both global and local learning while $\alpha=0.7$.
As reported in \reftable{abstudy}, we can draw the following conclusions:
First, compared with local learning only in \reftable{table:asm}, simultaneously optimizing global learning and local learning improve classification accuracy by 0.93\%.
Second, we first employed SA in \cite{Wang2018NonlocalNN} and MHSA in \cite{Transformer}. In the two works, the proposed attention module have four pointwise convolution layers, $Q_{1\times1}$, $K_{1\times1}$, $V_{1\times1}$, $O_{1\times1}$. 
However, this leads to over-fitting as there is a drop in accuracy from 32.78\% to 32.74\%.
After careful refinement, we find that $Q_{1\times1}$ and $O_{1\times1}$ are the two key pointwise convolution layers.
Third, the UpsampleAdd refers to $\rm \mathit{F}_{\mathit{l_s}} = \mathit{C^\prime}_{\mathit{l_s}}+Upsample_{2\times2}(\mathit{F}_{\mathit{l_s}+1})$.
Without UpsampleAdd, attended $\mathit{C^\prime}_{\mathit{l_s}}$ is directly sent to a local classifier for classification.
Although there is a drop in accuracy without UpsampleAdd, it shows the effectiveness of MHCCA in modeling CA, as the accuracy is 0.35\% higher compared with $Base$.
Forth, L-CAM is effective in highlighting level-wise spatial details. 
For pooling operations, $max$ selects the most important spatial positions among CAMs while $mean$ fuses all CAMs in a global view. 
The result shows the complementarity between the two pooling operations.
\begin{figure}[!t]
    \renewcommand\arraystretch{0.01}
	\centering
	\resizebox{\columnwidth}{!}{
		\begin{tabular}{c}
		\includegraphics[width=0.38\columnwidth]{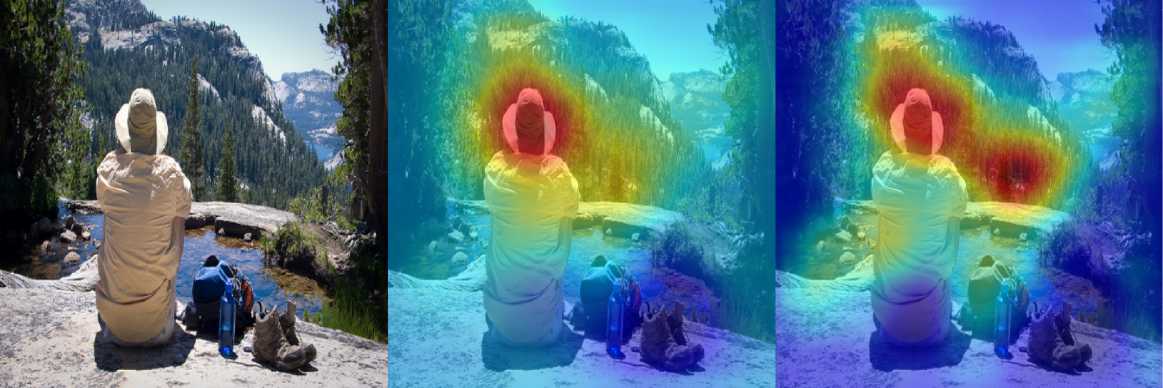}\\ \includegraphics[width=0.38\columnwidth]{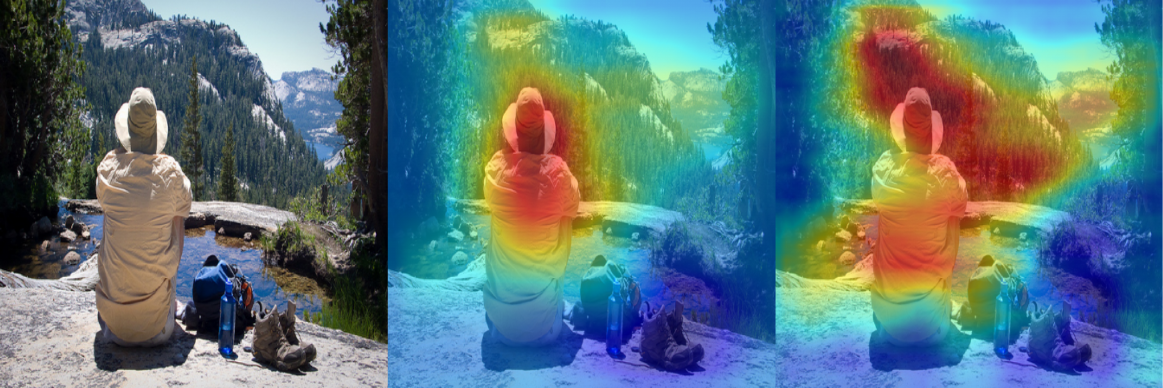}\\
		\end{tabular}}
		\caption{Visualization of spatial attention maps for emotion classification at $l_a=1$ (middle) and $l_a=2$ (right) without (upper) or with (lower) L-CAM. The original image from the FI dataset are in the left. The class of the sample image is contentment.}
		\label{fig:cam}

\end{figure}

\subsection{Visualization}
% We present some spatial attention visualization results \footnote{Please download the \href{https://www.dropbox.com/s/hjr4216175ayg4w/MDAN_CVPR_Supp.pdf?dl=0}{\underline{Suppl. Material}} for more attention visualization results of both L-CAM and MHCCA under different emotion hierarchies.} of the image in \reffig{fig:cam}.
%
We present some spatial attention visualization results \footnote{See \url{https://www.dropbox.com/s/hjr4216175ayg4w/MDAN_CVPR_Supp.pdf?dl=0} for more attention visualization results of both L-CAM and MHCCA under different emotion hierarchies.} of the image in \reffig{fig:cam}.
The attention map at $l_a=1$ considers highly semantic and salient objects which are discriminative enough to make a prediction at a low affective level.
However, as affective level increases, not only objects with strong semantic meaning but also spatial details such as background, color, texture details, are significant to predict emotions a higher affective level.
Horizontally, we can observe that salient object, the back view of man in a hat, was highlighted at $l_a=1$, after which the natural view with no salient object but with rich texture, color, and shape at the background was highlighted at $l_a=2$.
Vertically, more spatial details are highlighted with L-CAM, resulting in a more discriminative feature map.
This verifies that the affective semantic mapping and leveraging subordinate relationships in emotion hierarchies is effective.

\section{Conclusion}
The paper addresses the problem of fine-grained emotion classification by disentangling the affective gap into several smaller gaps. 
We present MDAN, a two-branch deep framework based on FPN.
The bottom-up branch learns globally and follows emotion hierarchies largely, while the top-down branch explores the level-wise discrimination through several local classifiers and propagates it forward to guide the classification at a higher affective level. 
Moreover, two attention modules: MHCCA and L-CAM are appended at each semantic level of MDAN to model level-wise channel dependencies and spatial attention.
Experimental results show the effectiveness of our method against the state-of-the-art on six benchmark datasets at all affective levels.

%%%%%%%%% REFERENCES
{\small
\bibliographystyle{ieee_fullname}
\bibliography{egbib.bib}

\begin{thebibliography}{10}\itemsep=-1pt

\bibitem{Borth2013SentiBankLO}
Damian Borth, Tao Chen, R. Ji, and Shih-Fu Chang.
\newblock Sentibank: large-scale ontology and classifiers for detecting
  sentiment and emotions in visual content.
\newblock In {\em MM '13}, 2013.

\bibitem{Carion2020EndtoEndOD}
Nicolas Carion, Francisco Massa, Gabriel Synnaeve, Nicolas Usunier, Alexander
  Kirillov, and Sergey Zagoruyko.
\newblock End-to-end object detection with transformers.
\newblock {\em ArXiv}, abs/2005.12872, 2020.

\bibitem{Chen2014DeepSentiBankVS}
Tao Chen, Damian Borth, Trevor Darrell, and Shih-Fu Chang.
\newblock Deepsentibank: Visual sentiment concept classification with deep
  convolutional neural networks.
\newblock {\em ArXiv}, abs/1410.8586, 2014.

\bibitem{Dosovitskiy2021AnII}
A. Dosovitskiy, L. Beyer, Alexander Kolesnikov, Dirk Weissenborn, Xiaohua Zhai,
  Thomas Unterthiner, M. Dehghani, Matthias Minderer, G. Heigold, S. Gelly,
  Jakob Uszkoreit, and N. Houlsby.
\newblock An image is worth 16x16 words: Transformers for image recognition at
  scale.
\newblock {\em ArXiv}, abs/2010.11929, 2021.

\bibitem{ekman}
Paul Ekman.
\newblock An argument for basic emotions.
\newblock {\em Cognition and Emotion}, 6(3-4):169--200, 1992.

\bibitem{He2016DeepRL}
Kaiming He, X. Zhang, Shaoqing Ren, and Jian Sun.
\newblock Deep residual learning for image recognition.
\newblock {\em 2016 IEEE Conference on Computer Vision and Pattern Recognition
  (CVPR)}, pages 770--778, 2016.

\bibitem{Krizhevsky2012ImageNetCW}
A. Krizhevsky, Ilya Sutskever, and Geoffrey~E. Hinton.
\newblock Imagenet classification with deep convolutional neural networks.
\newblock {\em Communications of the ACM}, 60:84 -- 90, 2012.

\bibitem{Lin2017FeaturePN}
Tsung-Yi Lin, Piotr Doll{\'a}r, Ross~B. Girshick, Kaiming He, Bharath
  Hariharan, and Serge~J. Belongie.
\newblock Feature pyramid networks for object detection.
\newblock {\em 2017 IEEE Conference on Computer Vision and Pattern Recognition
  (CVPR)}, pages 936--944, 2017.

\bibitem{lu2012shape}
Xin Lu, Poonam Suryanarayan, Reginald~B Adams~Jr, Jia Li, Michelle~G Newman,
  and James~Z Wang.
\newblock On shape and the computability of emotions.
\newblock In {\em Proceedings of the 20th ACM international conference on
  Multimedia}, pages 229--238, 2012.

\bibitem{Machajdik2010AffectiveIC}
J. Machajdik and A. Hanbury.
\newblock Affective image classification using features inspired by psychology
  and art theory.
\newblock {\em Proceedings of the 18th ACM international conference on
  Multimedia}, 2010.

\bibitem{Mikels2005EmotionalCD}
Joseph~A. Mikels, B. Fredrickson, G.~R. Larkin, Casey~M. Lindberg, Sam~J.
  Maglio, and P. Reuter-Lorenz.
\newblock Emotional category data on images from the international affective
  picture system.
\newblock {\em Behavior Research Methods}, 37:626--630, 2005.

\bibitem{Panda2018ContemplatingVE}
R. Panda, Jianming Zhang, Haoxiang Li, Joon-Young Lee, X. Lu, and A.
  Roy-Chowdhury.
\newblock Contemplating visual emotions: Understanding and overcoming dataset
  bias.
\newblock In {\em ECCV}, 2018.

\bibitem{parrott2001emotions}
W~Gerrod Parrott.
\newblock {\em Emotions in social psychology: Essential readings}.
\newblock psychology press, 2001.

\bibitem{rao2019multi}
Tianrong Rao, Xiaoxu Li, Haimin Zhang, and Min Xu.
\newblock Multi-level region-based convolutional neural network for image
  emotion classification.
\newblock {\em Neurocomputing}, 333:429--439, 2019.

\bibitem{Sartori2015WhosAO}
Andreza Sartori, Dubravko Culibrk, Yan Yan, and N. Sebe.
\newblock Who's afraid of itten: Using the art theory of color combination to
  analyze emotions in abstract paintings.
\newblock {\em Proceedings of the 23rd ACM international conference on
  Multimedia}, 2015.

\bibitem{She2020WSCNetWS}
Dongyu She, Jufeng Yang, Ming-Ming Cheng, Yu-Kun Lai, Paul~L. Rosin, and Liang
  Wang.
\newblock Wscnet: Weakly supervised coupled networks for visual sentiment
  classification and detection.
\newblock {\em IEEE Transactions on Multimedia}, 22:1358--1371, 2020.

\bibitem{Simonyan2015VeryDC}
K. Simonyan and Andrew Zisserman.
\newblock Very deep convolutional networks for large-scale image recognition.
\newblock {\em CoRR}, abs/1409.1556, 2015.

\bibitem{Srinivas2021BottleneckTF}
A. Srinivas, Tsung-Yi Lin, Niki Parmar, Jonathon Shlens, P. Abbeel, and Ashish
  Vaswani.
\newblock Bottleneck transformers for visual recognition.
\newblock {\em ArXiv}, abs/2101.11605, 2021.

\bibitem{Strudel2021SegmenterTF}
Robin A.~M. Strudel, Ricardo~Garcia Pinel, Ivan Laptev, and Cordelia Schmid.
\newblock Segmenter: Transformer for semantic segmentation.
\newblock {\em ArXiv}, abs/2105.05633, 2021.

\bibitem{Sun2016DiscoveringAR}
Ming Sun, Jufeng Yang, Kai Wang, and Hui Shen.
\newblock Discovering affective regions in deep convolutional neural networks
  for visual sentiment prediction.
\newblock {\em 2016 IEEE International Conference on Multimedia and Expo
  (ICME)}, pages 1--6, 2016.

\bibitem{Transformer}
Ashish Vaswani, Noam Shazeer, Niki Parmar, Jakob Uszkoreit, Llion Jones,
  Aidan~N Gomez, \L~ukasz Kaiser, and Illia Polosukhin.
\newblock Attention is all you need.
\newblock In I. Guyon, U.~V. Luxburg, S. Bengio, H. Wallach, R. Fergus, S.
  Vishwanathan, and R. Garnett, editors, {\em Advances in Neural Information
  Processing Systems}, volume~30. Curran Associates, Inc., 2017.

\bibitem{Wang2018NonlocalNN}
X. Wang, Ross~B. Girshick, A. Gupta, and Kaiming He.
\newblock Non-local neural networks.
\newblock {\em 2018 IEEE/CVF Conference on Computer Vision and Pattern
  Recognition}, pages 7794--7803, 2018.

\bibitem{yang2018auto}
Jufeng Yang, Dongyu She, Ming Sun, Ming-Ming Cheng, Paul~L. Rosin, and Liang
  Wang.
\newblock Visual sentiment prediction based on automatic discovery of affective
  regions.
\newblock {\em IEEE Transactions on Multimedia}, 20(9):2513--2525, 2018.

\bibitem{You2017VisualSA}
Quanzeng You, Hailin Jin, and Jiebo Luo.
\newblock Visual sentiment analysis by attending on local image regions.
\newblock In {\em AAAI}, 2017.

\bibitem{You2015RobustIS}
Quanzeng You, Jiebo Luo, Hailin Jin, and Jianchao Yang.
\newblock Robust image sentiment analysis using progressively trained and
  domain transferred deep networks.
\newblock In {\em AAAI}, 2015.

\bibitem{You2016BuildingAL}
Quanzeng You, Jiebo Luo, Hailin Jin, and Jianchao Yang.
\newblock Building a large scale dataset for image emotion recognition: The
  fine print and the benchmark.
\newblock In {\em AAAI}, 2016.

\bibitem{You2016CrossmodalityCR}
Quanzeng You, Jiebo Luo, Hailin Jin, and Jianchao Yang.
\newblock Cross-modality consistent regression for joint visual-textual
  sentiment analysis of social multimedia.
\newblock {\em Proceedings of the Ninth ACM International Conference on Web
  Search and Data Mining}, 2016.

\bibitem{Zhang2020WeaklySE}
Haimin Zhang and Min Xu.
\newblock Weakly supervised emotion intensity prediction for recognition of
  emotions in images.
\newblock {\em IEEE Transactions on Multimedia}, pages 1--1, 2020.

\bibitem{zhao2014exploring}
Sicheng Zhao, Yue Gao, Xiaolei Jiang, Hongxun Yao, Tat-Seng Chua, and Xiaoshuai
  Sun.
\newblock Exploring principles-of-art features for image emotion recognition.
\newblock In {\em Proceedings of the 22nd ACM international conference on
  Multimedia}, pages 47--56, 2014.

\bibitem{Zhao2019PDANetPD}
Sicheng Zhao, Zizhou Jia, H. Chen, L. Li, Guiguang Ding, and K. Keutzer.
\newblock Pdanet: Polarity-consistent deep attention network for fine-grained
  visual emotion regression.
\newblock {\em Proceedings of the 27th ACM International Conference on
  Multimedia}, 2019.

\bibitem{Zhao2016PredictingPE}
Sicheng Zhao, H. Yao, Yue Gao, R. Ji, Wenlong Xie, Xiaolei Jiang, and Tat-Seng
  Chua.
\newblock Predicting personalized emotion perceptions of social images.
\newblock {\em Proceedings of the 24th ACM international conference on
  Multimedia}, 2016.

\bibitem{Zhao2021AffectiveIC}
Sicheng Zhao, Xingxu Yao, Jufeng Yang, G. Jia, Guiguang Ding, Tat-Seng Chua, B.
  Schuller, and K. Keutzer.
\newblock Affective image content analysis: Two decades review and new
  perspectives.
\newblock {\em IEEE transactions on pattern analysis and machine intelligence},
  PP, 2021.

\bibitem{Zhou2016LearningDF}
B. Zhou, A. Khosla, {\`A}. Lapedriza, A. Oliva, and A. Torralba.
\newblock Learning deep features for discriminative localization.
\newblock {\em 2016 IEEE Conference on Computer Vision and Pattern Recognition
  (CVPR)}, pages 2921--2929, 2016.

\bibitem{rao2017dependency}
Xinge Zhu, Liang Li, Weigang Zhang, Tianrong Rao, Min Xu, Qingming Huang, and
  Dong Xu.
\newblock Dependency exploitation: A unified cnn-rnn approach for visual
  emotion recognition.
\newblock In {\em Proceedings of the Twenty-Sixth International Joint
  Conference on Artificial Intelligence, {IJCAI-17}}, pages 3595--3601, 2017.

\end{thebibliography}
}

\end{document}